\documentclass[10pt,twocolumn,letterpaper]{article}

\usepackage{iccv}
\usepackage{times}
\usepackage{epsfig}
\usepackage{graphicx}
\usepackage{amsmath}
\usepackage{amssymb}

\usepackage{booktabs}
\usepackage{bm}
\usepackage{multirow}
\usepackage{subfig}

\usepackage{amsmath}

\usepackage[dvipsnames]{xcolor}

\usepackage{algorithm}
\usepackage{algorithmic}
\usepackage[accsupp]{axessibility}  

\usepackage[breaklinks=true,bookmarks=false]{hyperref}

\usepackage[capitalize]{cleveref}
\crefname{section}{Sec.}{Secs.}
\Crefname{section}{Section}{Sections}
\Crefname{table}{Table}{Tables}
\crefname{table}{Tab.}{Tabs.}

\makeatletter
\renewcommand{\paragraph}{%
  \@startsection{paragraph}{4}%
  {\z@}{.2ex \@plus 1ex \@minus .2ex}{-1em}%
  {\normalfont\normalsize\bfseries}%
}
\makeatother

\iccvfinalcopy 

\def\iccvPaperID{9258} 

\ificcvfinal\fi

\begin{document}

\title{Beating Backdoor Attack at Its Own Game}


\author{
Min Liu$^{1}$,
Alberto Sangiovanni-Vincentelli$^2$,
Xiangyu Yue$^{3}$\\
$^{1}$Carnegie Mellon University,
$^{2}$UC Berkeley,
$^{3}$The Chinese University of Hong Kong\\
{\tt\small minliu2@cs.cmu.edu,
alberto@berkeley.edu,
xyyue@ie.cuhk.edu.hk}
}

\maketitle
\ificcvfinal\thispagestyle{empty}\fi

\begin{abstract}
Deep neural networks (DNNs) are vulnerable to backdoor attack, which does not affect the network's performance on clean data but would manipulate the network behavior once a trigger pattern is added. 
Existing defense methods have greatly reduced attack success rate, but their prediction accuracy on clean data still lags behind a clean model by a large margin. 
Inspired by the stealthiness and effectiveness of backdoor attack, we propose a simple but highly effective defense framework which injects non-adversarial backdoors targeting poisoned samples.
Following the general steps in backdoor attack, we detect a small set of suspected samples and then apply a poisoning strategy to them.
The non-adversarial backdoor, once triggered, suppresses the attacker's backdoor on poisoned data, but has limited influence on clean data.
The defense can be carried out during data preprocessing, without any modification to the standard end-to-end training pipeline.
We conduct extensive experiments on multiple benchmarks with different architectures and representative attacks. 
Results demonstrate that our method achieves state-of-the-art defense effectiveness with by far the lowest performance drop on clean data. Considering the surprising defense ability displayed by our framework, we call for more attention to utilizing backdoor for backdoor defense. 
Code is available at \href{https://github.com/minliu01/non-adversarial\_backdoor}{\textcolor{VioletRed}{https://github.com/minliu01/non-adversarial\_backdoor}}.
\end{abstract}

\section{Introduction}
In recent years, deep neural networks (DNNs) have achieved impressive performance across tasks, such as object detection \cite{simonyan2014very, ren2015faster}, speech recognition \cite{xiong2016achieving, baevski2020wav2vec} and machine translation \cite{sutskever2014sequence, vaswani2017attention}. With the increasing usage of DNNs, security of neural networks has attracted a lot of attention. 
Studies have shown that DNNs are especially vulnerable to backdoor attack \cite{weng2020trade}, a variant of data poisoning which fools the model to establish a false correlation between inserted patterns and target classes. Specifically, the adversary injects a trigger pattern to a small proportion of the training data. A network trained on the poisoned data has normal behavior on benign data, but deviates from its expected output when the trigger pattern is implanted. 

\begin{figure}[t]
    \centering
    \setlength{\abovecaptionskip}{0.4em}
    \includegraphics[width=0.70\linewidth]{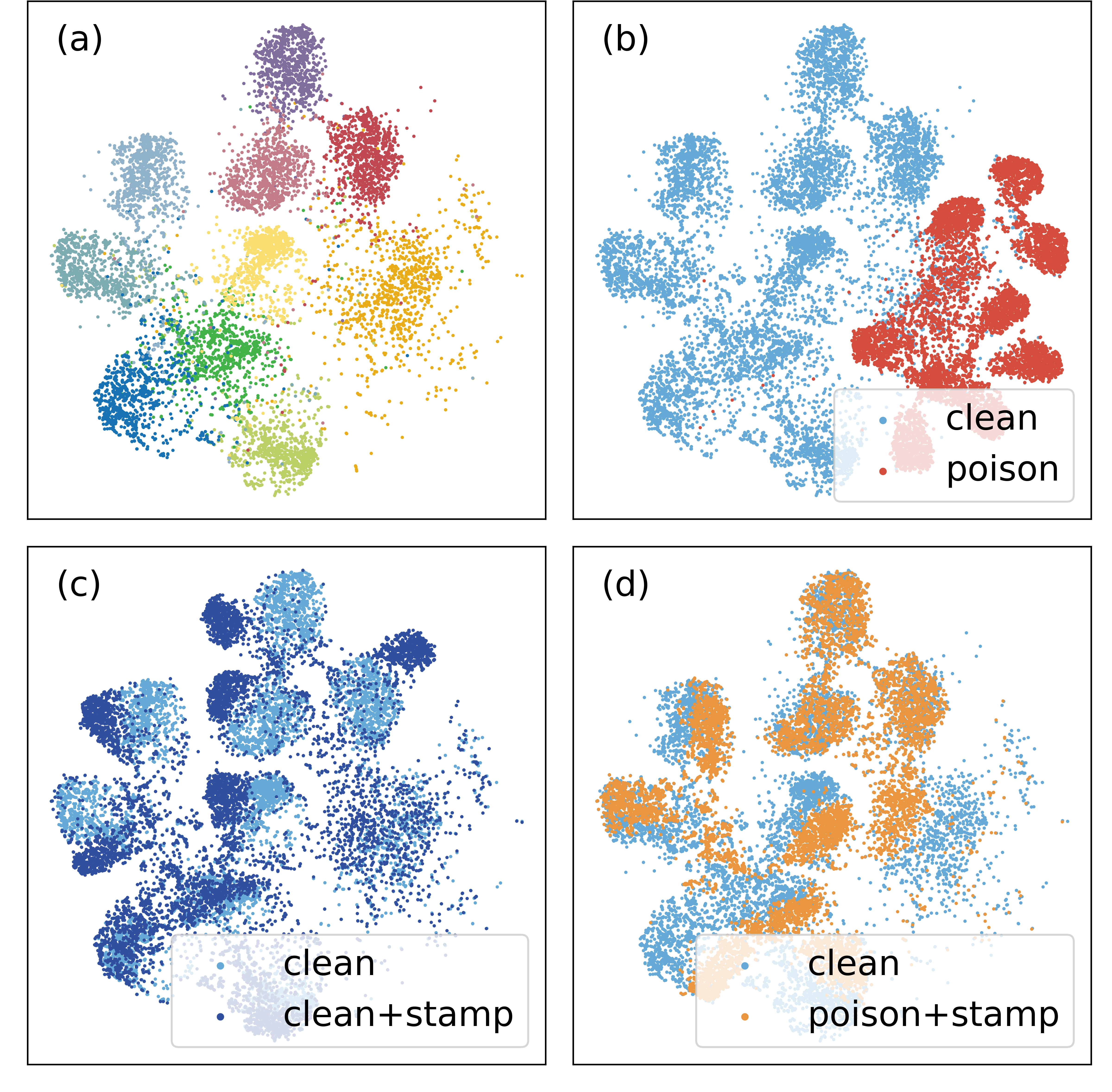}
    \caption{Representations under the effect of adversarial backdoor (AB) and non-adversarial backdoor (NAB), which are injected by attackers and defenders respectively.
    ``Stamp'' is the trigger pattern for NAB.
    (a) Clean samples are not influenced by backdoor.
    (b) AB changes model behavior on poisoned samples.
    (c) NAB is not triggered on clean samples.
    (d) NAB suppresses the effectiveness of AB on poisoned samples.
    }
    \label{fig:representations}
\end{figure}

To ensure the security of DNN systems, a lot of novel defense methods have been proposed in the past few years. Most of the defense methods try to either 1) avoid learning the backdoor during training or 2) erase it from a poisoned model at the end. 
Following idea 1), some studies detect and filter poisoned samples \cite{tang2021demon, chou2020sentinet}. Since a small number of poisoned samples slipping from detection can lead to a successful attack, simply filtering the potentially poisoned samples is not enough in most cases. A more realistic way is to adopt data separation as an intermediate procedure \cite{li2021anti, huang2022backdoor}.
Some other works pre-process the input to depress the effectiveness of injected patterns \cite{qiu2021deepsweep, hong2020effectiveness}. However, these methods have limited effects under the increasingly diverse attacking strategies.
Another line of work follows idea 2) \cite{li2021neural, wu2021adversarial}. Despite promising defense effectiveness, erasing-based methods suffer from performance drop due to the additional erasing stage. Performance on clean data still lags behind a clean model by a large margin.
Reducing the performance gap on clean data while maintaining satisfying defense effectiveness remains a challenging problem.

Under backdoor attack, representations of poisoned samples are dominated by the trigger pattern as shown in \cref{fig:representations}. Therefore, injecting the pattern can force a poisoned model to behave in a way expected by the attacker. 
Considering the effectiveness of such strategies, a natural question is whether backdoor can be utilized for defense purpose, that is to say \textit{beating backdoor attack at its own game}. 
To be more specific, a model might misbehave when only the trigger pattern is exposed, but the mishehavior should be suppressed once a benign pattern, which is called a \textit{stamp} in this paper, is injected to the poisoned sample.
There are three advantages behind this idea. \textit{First}, the defender only needs a small set of poisoned training samples to inject a backdoor, which is a much easier requirement than filtering all the poisoned data.
\textit{Second}, a backdoor targeting poisoned data, ideally, will not influence the model performance on clean data.
\textit{Finally}, the backdoor can be injected during data pre-processing, without any modification to the standard end-to-end training pipeline.

In this work, we propose a novel defense framework, \textit{Non-adversarial Backdoor (NAB)}, which suppresses backdoor attack by injecting a backdoor targeting poisoned samples.
Specifically, we first detect a small set of suspected samples using existing methods such as \cite{li2021anti, huang2022backdoor, hayase2021spectre}. 
Then we process these samples with a poisoning strategy, which consists of a stamping and a relabeling function. A pseudo label is generated for each detected sample and we stamp the samples with inconsistent orginal and pseudo labels.
In this way, we insert a non-adversarial backdoor which, once triggered, is expected to change model behaviors on poisoned data. 
Furthermore, NAB can be augmented with an efficient test data filtering technique by comparing the predictions with or without the stamp, ensuring the performance on poisoned data.
We instantiated the NAB framework and conducted experiments on CIFAR-10 \cite{krizhevsky2009learning} and tiny-ImageNet \cite{le2015tiny} over several representative backdoor attacks. Experiment results show that the method achieves state-of-the-art performance in both clean accuracy and defense effectiveness.
Extensive analyses demonstrate how NAB takes effect under different scenarios.

Our main contributions can be summarized as follows:
\begin{itemize}
\setlength\itemsep{-0.2em}
    \item We propose the idea of backdooring poisoned samples to suppress backdoor attack. To the best of our knowledge, our work is the first to utilize non-adversarial backdoor in backdoor defense.
    \item We transform the idea into a simple, flexible and effective defense framework, which can be easily augmented with a test data filtering technique.
    \item Extensive experiments are conducted and our method achieves state-of-the-art defense effectiveness with by far the lowest performance drop on clean data.
    
\end{itemize}

\section{Related Work}
\paragraph{Backdoor Attack.}
Backdoor attack is a type of attack involved in the training of DNNs, with the interesting property that the model works well on clean data but generates unexpected outputs once the attack is triggered. 
A main track of the attacks focuses on poisoning training data in an increasingly stealthier and more effective way by developing novel trigger patterns \cite{kaviani2021defense}. 
Attack methods for visual models, the mainstream of backdoor attack research, can be divided according to the visibility of patterns. 
Visible attacks inject human perceptible patterns like a single pixel \cite{tran2018spectral}, an explicit patch \cite{gu2017badnets, liu2017trojaning}, sample-specific patterns \cite{nguyen2020input}, or more complex and indistinguishable patterns like blending random noise \cite{chen2017targeted} and sinusoidal strips \cite{barni2019new}. 
Invisible attacks \cite{zhao2020clean, turner2018clean, saha2020hidden, li2020invisible, nguyen2021wanet, li2021invisible} are even more stealthy to human observers. 
Backdoor attacks can also be categorized into dirty-label attacks \cite{gu2017badnets, liu2017trojaning, nguyen2020input} and clean-label attacks \cite{barni2019new, turner2018clean}. Clean-label attacks are more difficult to detect since there lacks an obvious mismatch between the images and labels.
We also notice some non-poisoning based methods which induce backdoor by modifying other training settings \cite{li2018hu, li2021deeppayload} or the model weights \cite{clements2018backdoor, dumford2020backdooring, rakin2020tbt}.

\paragraph{Backdoor Defense.}
Existing backdoor defense methods aim to avoid learning the backdoor during training or erase the backdoor at the end. 
To avoid injection of the backdoor, various techniques detecting poisoned data have been proposed \cite{tran2018spectral, chen2018detecting, tang2021demon, chou2020sentinet, hayase2021spectre}. These methods alone cannot achieve successful defense when a fraction of poisoned samples escape from the detection. Instead of simply filtering all the poisoned samples, a more practical idea is to adopt data separation as an intermediate procedure. 
Some other works attempt to bypass the backdoor by pre-processing the input before passing it into the model \cite{qiu2021deepsweep, du2019robust, hong2020effectiveness}, but these methods typically have limited effects over the increasingly various attacks. 
Meanwhile, erasing methods try to mitigate the effects of backdoor after the model gets attacked \cite{li2021anti, zhao2020bridging, li2021neural, wu2021adversarial}. \cite{li2021anti} reduced attack success rate to a negligible level under several attacks, but the prediction accuracy on clean data still lags behind a well-trained clean model by a large margin. 
Our method adopts a data separation stage as in \cite{li2021anti, huang2022backdoor}. Nevertheless, the core idea, injecting a backdoor for defense purpose, is similar to none of the previous defense methods.

\begin{figure*}[t]
    \centering
    \includegraphics[width=0.95\linewidth]{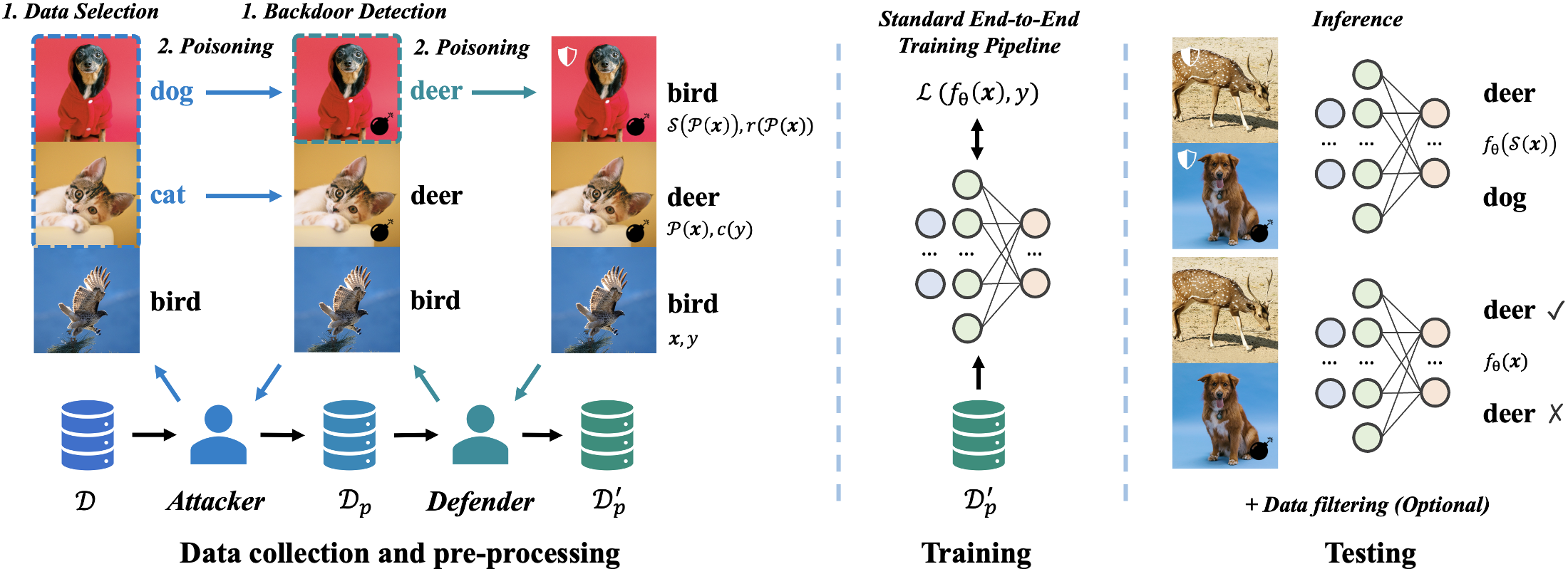} 
    \setlength{\abovecaptionskip}{0.2em}
    \caption{Overview of the proposed framework. 
    The attacker injects an adversarial backdoor by selecting and poisoning a set of clean samples.
    After obtaining the dataset, the defender detect and poison a set of suspected samples to inject the non-adversarial backdoor.
    Both attack and defense take place in the standard end-to-end training pipeline.
    In the testing stage, we stamp each input to keep the non-adversarial backdoor triggered.
    We can also adopt a test data filtering technique by comparing the predictions with or without the stamp. Samples with inconsistent predictions are identified as poisoned.}
    \label{fig:overview}
    \vspace{-0.6em}
\end{figure*}

\paragraph{Non-Adversarial Backdoor.}
Non-adversarial applications of backdoor has been proposed before, including watermark-based authentication \cite{adi2018turning}, protection of open-sourced datasets \cite{li2020open} and neural networks interpretability \cite{zhao2021deep}. \cite{shan2021using} also injected a backdoor to hide weaknesses in a model under adversarial attack \cite{moosavi2017universal}.
However, our work is the first attempt to utilize backdoor in defense against backdoor attack.

\section{Method}
\subsection{Preliminary}
\paragraph{Threat Model.} In this paper, we assume that the attacker has full control over the data source, is capable of arbitrarily changing the images and relabeling them with target classes, but does not have access to the model and training process. The defender has control over the model, training process, and data once obtained from the data source, but does not know the proportion and distribution of the poisoned samples, target classes, and attacking strategies. Given some partially poisoned data, the defender aims to train a model that preserves accuracy on clean data and avoids predicting the target class on poisoned data.

\paragraph{Backdoor Attack.} The attacker first obtains a clean training set $\mathcal{D}=\{(\bm{x}_i, y_i)\}_{i=1}^N$, where $\bm{x}_i \in \mathcal{X}$ is an image and $y_i \in \mathcal{Y}$ is the corresponding label. The poisoning strategy consists of two parts: $\mathcal{P}:\mathcal{X}\rightarrow \mathcal{X}$ applies a trigger pattern to the image and $c:\mathcal{Y}\rightarrow\mathcal{Y}$ replaces the label with target label. The attacker selects a subset of clean data and generates a set of malign samples $\mathcal{D}_m=\{(\mathcal{P}(\bm{x}), c(y))\}$ accordingly using the poisoning strategy, where $\lambda={|\mathcal{D}_{m}|} / {|\mathcal{D}|}$ is the poisoning rate. Merging $\mathcal{D}_m$ with the remaining clean training data, the attacker generates a poisoned dataset $\mathcal{D}_p$ and releases it to potential victims.

The empirical error on a poisoned dataset can be decomposed into a \textit{clean loss} and an \textit{attack loss} \cite{li2021anti}:
\begin{equation}
\label{eq:two_tasks}
    \mathbb{E}_{(\bm{x}, y)\sim \mathcal{D}}[\ell(f_\theta(\bm{x}), y)]+\mathbb{E}_{(\bm{x}, y)\sim \mathcal{D}_m}[\ell(f_\theta(\bm{x}), y)]
\end{equation}
where $\ell(\cdot)$ is the loss function and $f_\theta: \mathcal{X}\rightarrow\mathcal{Y}$ is the neural network. Minimizing the first term above encourages the model to learn the image classification task while the second one forces the learning of a correlation between the trigger pattern and target class. Both tasks can be learned well due to the excessive learning ability of neural networks \cite{li2022backdoor}, making backdoor attack effective and hard to detect.

\subsection{Non-Adversarial Backdoor}
The success of backdoor attack leads us to think about the feasibility of utilizing backdoor for defense purpose. An attacker wants the model to classify a benign sample $\bm{x}$ to the target class. In the same way, a defender wants the model to classify a poisoned sample $\mathcal{P}(\bm{x})$ to any but the target class. The similarity between the objectives makes it a natural idea to apply backdoor in both attack and defense settings, while the latter was rarely explored.

Based on the idea above, we propose a defense framework \textit{Non-Adversarial Backdoor} (NAB). 
In this section, we assume that a small set of suspected samples $\mathcal{D}_s'\subset\mathcal{D}_p$ and a poisoning strategy are available. 
The defender's poisoning strategy also has two components: 1) $\mathcal{S}: \mathcal{X}\rightarrow\mathcal{X}$ applies a trigger pattern, which is called a \textit{stamp} to tell from the adversarial trigger pattern, and 2) $r: \mathcal{X}\rightarrow \mathcal{Y}$ generates a pseudo label conditioned on the image. 
Details of backdoor detection and poisoning strategies are discussed in \cref{bd_ps_sec}. 
We then generate a set of stamped samples $\mathcal{D}_m'=\{ (\mathcal{S}(\bm{x}), r(\bm{x})) | (\bm{x}, y)\in \mathcal{D}_s' \land r(\bm{x}) \neq y \}$. Note that the proportion of stamped samples is typically lower than the detection rate $\mu={|\mathcal{D}_s'|} / {|\mathcal{D}_p|}$, as we avoid stamping samples whose labels remain unchanged to let the backdoor take effect. Merging $\mathcal{D}_m'$ with data that are not stamped, we obtain the processed dataset $D_p'$ for training. The defense framework can be implemented during data preprocessing, without any modification to the end-to-end training pipeline. During inference, we stamp all inputs for defense.

In NAB, we further decompose the attack loss in \cref{eq:two_tasks} into the original attack loss and the \textit{defense loss}:
\begin{equation}
    \mathbb{E}_{(\bm{x}, y)\sim \mathcal{D}_m}[\ell(f_\theta(\bm{x}), y)] + \mathbb{E}_{(\bm{x}, y)\sim \mathcal{D}_m'}[\ell(f_\theta(\bm{x}), y)]
    \label{eq:defense_task}
\end{equation}
Jointly optimizing the model using the three losses leaves two backdoors in the network: an adversarial backdoor triggered by $\mathcal{P}(\cdot)$ and a non-adversarial backdoor triggered by $\mathcal{S}(\mathcal{P}(\cdot))$. The non-adversarial one prevents a poisoned sample with stamp from being classified to the target class. Typically, $D_{s}'$ is a mixture of poisoned and clean samples due to mistakes of detection methods. When the detection accuracy is low, the non-adversarial backdoor might influence the performance on clean data. Further analysis is presented in \cref{sec:further_ana}.

\subsection{Backdoor Detection and Poisoning Strategy}
\label{bd_ps_sec}
While attackers can select samples to poison randomly and simply label them with the target class, defenders need more deliberation on data selection and relabeling strategy.

\paragraph{Backdoor Detection.} To create a backdoor targeting poisoned data, we detect a set of suspicious samples $\mathcal{D}_s'$ from $\mathcal{D}_p$ with ratio $\mu$. \textit{Detection accuracy} is the ratio of poisoned samples in $\mathcal{D}_s'$. Different from detection-based defenses that aims at filtering all the poisoned samples, $\mu$ is typically smaller than the poisoning rate $\lambda$ in NAB as we only need part of the poisoned data for backdoor injection. 

\paragraph{Poisoning Strategy.} The stamping function $\mathcal{S}(\cdot)$ is less important as long as it is perceptible to neural networks. We care more about the relabeling function $r(\cdot)$ which generates pseudo labels to approximate true labels. 
Although randomly generated pseudo labels suffices to create the non-adversarial backdoor, a higher \textit{pseudo label accuracy} can help preserve the performance on clean data when the detection method malfunctions.

Many existing or naive methods can fulfill the relabeling and simplified backdoor detection tasks effectively \cite{hayase2021spectre, li2021anti, huang2022backdoor}. Nevertheless, the NAB framework is independent of any specific detection method or poisoning strategy. As the chasing game between backdoor attack and defense goes on, stronger methods are likely to show up in the future, and NAB can be easily instantiated with the latest techniques. The flexibility and portability ensures the long-term value of our framework.

\subsection{Test Data filtering}
A wide range of previous works consider minimizing the attack success rate as their only goal on poisoned samples. However, misclassification might still happen even if the image is not classified to the target class, bringing about unintended consequences. If we add a requirement to the threat model that all poisoned test samples should be either identified or correctly classified, the defense effectiveness of some existing methods will be less satisfying.

An additional benefit of NAB is that it can be easily augmented with a test data filtering technique. Ideally, the prediction results on a clean sample $\bm{x}$ and its stamped version $\mathcal{S}(\bm{x})$ are both the true label $y$. However, the model tends to predict $c(y)$ on $\mathcal{P}(\bm{x})$ and $r(\mathcal{P}(\bm{x}))$ on $\mathcal{S}(\mathcal{P}(\bm{x}))$, which are expected to be different due to the defender's backdoor. Based on the observation above, we identify samples with $f_\theta(\bm{x})\neq f_\theta(\mathcal{S}(\bm{x}))$ as poisoned and reject them during inference. In this way, the augmented NAB can handle poisoned data appropriately with a high accuracy.

\section{Experiments}
\subsection{Experiment Settings}
\paragraph{Attack.} Experiments are conducted under 5 representative backdoor attacks, including two classical attacks: BadNets attack (patch-based) \cite{gu2017badnets} and Blend attack (blending-based) \cite{chen2017targeted}, two advanced attacks: Dynamic attack (sample-specific) \cite{nguyen2020input} and WaNet attack (invisible) \cite{nguyen2021wanet}, and one label-consistent attack: Clean-Label attack \cite{turner2018clean}. 
We follow the configurations suggested in the original papers, including the trigger patterns and trigger sizes. 
Performance of the attacks are evaluated on two datasets: CIFAR-10 (10 classes, 50k samples) \cite{krizhevsky2009learning} and tiny-ImageNet (200 classes, 100k samples) \cite{le2015tiny}. 
Dynamic attack and Clean-Label attack (CL) are omitted on tiny-ImageNet for failure of reproduction.
The target label is set to 0 for both datasets. We set the poisoning rate $\lambda=0.1$ for the first four attacks, and $\lambda=0.25$ (2.5\% of the whole training set) for CL.

\paragraph{Defense and Training.} 
We instantiate the NAB framework with 3 backdoor detection techniques and 2 relabeling strategies throughout our experiments. The detection rate $\mu$ is set to $0.05$ for the following methods:
\vspace{-0.4em}
\begin{itemize}
\setlength\itemsep{-0.4em}
    \item \textbf{Local Gradient Ascent (LGA)} \cite{li2021anti}: Train with a tailored loss function in early epochs and isolate samples with lower training losses.
    \item \textbf{Label-Noise Learning (LN)} \cite{huang2022backdoor}: Train a classier appended to a self-supervised learning (SSL) pertained feature extractor with label-noise learning \cite{wang2019symmetric}, and capture low-credible samples.
    \item \textbf{SPECTRE} \cite{hayase2021spectre}: Use robust covariance estimation to amplify the spectral signature of poisoned data and detect them with QUantum Entropy (QUE) scores.
\end{itemize}
\vspace{-0.4em}
In our poisoning strategy, $\mathcal{S}(\cdot)$ simply applies a $2\times 2$ patch with value 0 on the upper left corner of the samples. We adopt the following strategies for pseudo label generation:
\vspace{-0.4em}
\begin{itemize}
\setlength\itemsep{-0.4em}
    \item \textbf{Verified Data (VD)}: Train a label predictor with supervised learning on a small collection of verified data (5\% of the training set as assumed in \cite{li2021neural}).
    \item \textbf{Nearest-Center (NC)}: Obtain representations using a SSL-pretrained model and assign pseudo labels according to the nearest center.
\end{itemize}
\vspace{-0.4em}
We \textit{do not assume} that the methods above are state-of-the-art. They are chosen for their simplicity and can be safely replaced with comparable or stronger methods. LN and NC are introduced because one of our baselines \cite{huang2022backdoor} relies on a SSL stage. Experiments are conducted on ResNet-18 (by default) and ResNet-50 \cite{he2016deep}. We train the models for 100 epochs with three data augmentations: random crop, horizontal flipping and cutout. The optimizer is Stochastic Gradient Descent (SGD) with momentum 0.9. Learning rate is set to 0.1 and decays with the cosine decay schedule \cite{loshchilov2016sgdr}. More details are presented in the supplementary material.

\paragraph{Baselines.} 
We compare our method with 3 state-of-the-art defense methods: 
1) Neural Attention Distillation (NAD) \cite{li2021neural} uses 5\% of clean training data to fine-tune a student network under the guidance of a teacher model. We use the same set of verified data for NAD and the relabeling strategy VD.
2) Anti-Backdoor Learning (ABL) \cite{li2021anti} unlearns the backdoor using a small set of isolated data. Note that an additional fine-tuning stage is added before backdoor unlearning to improve clean accuracy for fair comparison. 
3) Decoupling-Based backdoor Defense (DBD) \cite{huang2022backdoor} divides the training pipeline into a three stages to prevent learning the backdoor. Despite its impressive performance under some attacks, we find that DBD fails when the poisoned samples are clustered after the self-supervised learning stage under some attacks (\textit{e.g.} Dynamic attack \cite{nguyen2020input}). Besides, DBD was tested without applying trigger patterns to the target class, but its performance drops when the constraint is removed. We leave a detailed discussion of the weaknesses in the supplementary material, and provide a separate comparison with our method following their original settings except for the poisoning rate.

\paragraph{Metrics.} 
We adopt two widely used metrics for the main results: attack success rate (ASR, ratio of poisoned samples mistakenly classified to the target class) and Clean Accuracy (CA, ratio of correctly predicted clean samples). 
To test the effectiveness of our data filtering method, we further introduce backdoor accuracy (BA, ratio of correctly predicted backdoor samples), ratio of rejected clean data (C-REJ), prediction success rate (PSR, ratio of correctly predicted \textit{and not} rejected clean samples), ratio of rejected poisoned data (B-REJ), and defense success rate (DSR, ratio of correctly predicted \textit{or} rejected poisoned samples)

\begin{table*}
    \centering
    \setlength\tabcolsep{3pt}
    \renewcommand{\arraystretch}{1.15}
    \begin{tabular}{c|cccccccc|cccccccc}
    \toprule[1.2pt]
        \textbf{Arch} & \multicolumn{8}{c|}{\textbf{ResNet-18}} & \multicolumn{8}{c}{\textbf{ResNet-50}} \\\hline
        {\textbf{Defense}} & \multicolumn{2}{c}{\textbf{No Defense}} & \multicolumn{2}{c}{\textbf{NAD}} & \multicolumn{2}{c}{\textbf{ABL}} & \multicolumn{2}{c|}{\textbf{NAB (Ours)}} & \multicolumn{2}{c}{\textbf{No Defense}} & \multicolumn{2}{c}{\textbf{NAD}} & \multicolumn{2}{c}{\textbf{ABL}} & \multicolumn{2}{c}{\textbf{NAB (Ours)}} \\\hline
         \textbf{Attack $\downarrow$} & CA & ASR & CA & ASR & CA & ASR & CA & ASR & CA & ASR & CA & ASR & CA & ASR & CA & ASR \\\hline
        BadNets & 93.99 & 100 & 89.09 & 2.04 & 91.85 & \textbf{0.26} & \textbf{93.26} & 0.93 & 94.09 & 99.52 & 89.97 & 1.29 & 92.80 & 0.50 & \textbf{93.44} & \textbf{0.00} \\
        Blend & 94.09 & 100 & 89.29 & 1.22 & 89.87 & 1.62 & \textbf{93.18} & \textbf{0.29} & 94.26 & 99.98 & 90.04 & 1.03 & 88.11 & 1.41 & \textbf{94.34} & \textbf{0.09} \\
        Dynamic & 94.29 & 99.99 & 89.11 & 10.28 & 91.64 & 1.74 & \textbf{93.75} & \textbf{0.24} & 94.00 & 99.98 & 89.80 & 4.53 & 92.50 & 1.30 & \textbf{94.23} & \textbf{0.12} \\
        WaNet & 93.06 & 97.53 & 88.52 & 1.31 & 89.57 & 9.11 & \textbf{90.36} & \textbf{0.67} & 93.19 & 97.02 & 89.90 & 1.64 & 88.39 & 4.11 & \textbf{91.54} & \textbf{0.34} \\
        CL & 94.66 & 99.73 & 88.97 & 4.63 & 87.27 & 0.61 & \textbf{91.63} & \textbf{0.48} & 94.70 & 91.58 & 90.16 & 4.22 & 88.24 & 0.98 & \textbf{91.50} & \textbf{0.40} \\\hline
        \textbf{Average} & 94.02 & 99.45 & 89.00 & 3.90 & 90.04 & 2.67 & \textbf{92.44} & \textbf{0.52} & 94.05 & 97.62 & 89.97 & 2.54 & 90.00 & 1.66 & \textbf{93.01} & \textbf{0.19} \\
    \bottomrule[1.2pt]
    \end{tabular}
    \caption{Attack success rate (\%) and clean accuracy (\%) of NAD, ABL and our proposed method against 5 attacks over ResNet-18 and ResNet-50. The benchmark is CIFAR-10. We bold the best defense results under each attack.}
    \label{tab:main_cifar10}
\end{table*}

\begin{table}
    \centering
    \setlength\tabcolsep{1.6pt}
    \renewcommand{\arraystretch}{1.15}
    \begin{tabular}{c|cc|cc|cc|cc}
    \toprule[1.2pt]
        {\textbf{Defense}} & \multicolumn{2}{c|}{\textbf{No Defense}} & \multicolumn{2}{c|}{\textbf{NAD}} & \multicolumn{2}{c|}{\textbf{ABL}} & \multicolumn{2}{c}{\textbf{NAB (Ours)}} \\\hline
        \textbf{Attack $\downarrow$} & CA & ASR & CA & ASR & CA & ASR & CA & ASR  \\\hline
        BadNets & 64.85 & 99.94 & 61.80 & 1.58 & 62.21 & \textbf{0.00} & \textbf{63.14} & 0.89 \\
        Blend & 64.07 & 99.03 & 60.06 & 7.25 & 56.54 & \textbf{0.00} & \textbf{63.04} & 1.09 \\
        WaNet & 64.33 & 99.86 & 60.16 & 3.91 & 55.79 & 0.69 & \textbf{62.27} & \textbf{0.63} \\\hline
        \textbf{Average} & 64.42 & 99.61 & 60.67 & 4.25 & 58.18 & \textbf{0.23} & \textbf{62.81} & 0.87 \\
        
    \bottomrule[1.2pt]
    \end{tabular}
    \caption{Defense effectiveness (\%) of baselines and our method against 3 attacks on tiny-ImageNet.}
    \label{tab:main_tiny}
\end{table}

\begin{table}
    \centering
    \setlength\tabcolsep{3.8pt}
    \renewcommand{\arraystretch}{1.15}
    \begin{tabular}{c|cc|cc|cc}
    \toprule[1.2pt]
        \textbf{Defense} & \multicolumn{2}{c|}{\textbf{DBD}} & \multicolumn{2}{c|}{\textbf{NAB (Ours)}} & \multicolumn{2}{c}{\textbf{NAB* (Ours)}} \\\hline
        \textbf{Attack $\downarrow$} & CA & ASR & CA & ASR & CA & ASR  \\\hline
        BadNets & 92.60 & 1.49 & 93.69 & \textbf{0.33} & \textbf{94.44} & 0.42 \\
        Blend & 92.64 & 1.87 & 94.18 & 0.48 & \textbf{94.85} & \textbf{0.46} \\
        WaNet & 90.71 & 1.04 & 92.83 & \textbf{0.54} & \textbf{93.55} & 0.66 \\
        CL    & 92.94 & 0.95 & 93.83 & 1.31 & \textbf{94.57} & \textbf{0.71} \\\hline
        \textbf{Average} & 92.22 & 1.34 & 93.63 & 0.67 & \textbf{94.35} & \textbf{0.56}\\
    \bottomrule[1.2pt]
    \end{tabular}
    \caption{Defense effectiveness (\%) of self-supervised learning based defense methods on CIFAR-10.}
    \label{tab:main_ssl}
\end{table}

\begin{table}
    \centering
    \setlength\tabcolsep{2pt}
    \renewcommand{\arraystretch}{1.15}
    \begin{tabular}{c|ccc|cccc}
    \toprule[1.2pt]
        {\textbf{Defense}} & \textbf{NAD} & \textbf{ABL} & \textbf{NAB} & \multicolumn{4}{c}{\textbf{NAB + filtering}} \\\hline
        \textbf{Attack $\downarrow$} & \multicolumn{3}{c|}{BA} & C-REJ & PSR & B-REJ & DSR  \\\hline
        BadNets & 87.66 & 91.69 & 72.10 & 2.83 & 92.49 & 98.61 & {99.14} \\
        Blend & 85.26 & 89.64 & 72.47 & 1.54 & 92.79 & 97.17 & {99.68} \\
        Dynamic & 66.75 & 85.47 & 65.38 & 1.71 & 93.26 & 96.94 & {99.67} \\
        WaNet & 86.51 & 81.48 & 79.51 & 6.47 & 89.04 & 89.30 & {99.15} \\
        CL & 86.06 & 86.02 & 75.28 & 4.41 & 90.95 & 89.83 & {99.33} \\\hline
        \textbf{Average} & 82.45 & 86.86 & 72.95 & 3.39 & 91.71 & 94.37 & {99.39} \\
    \bottomrule[1.2pt]
    \end{tabular}
    \caption{Backdoor accuracy (\%) and effectiveness (\%) of data filtering on CIFAR-10, ResNet-18.}
    \label{tab:filter}
\end{table}

\begin{figure} [t!]
\centering
\setlength{\abovecaptionskip}{0.2em}
    \includegraphics[width=0.95\linewidth]
    {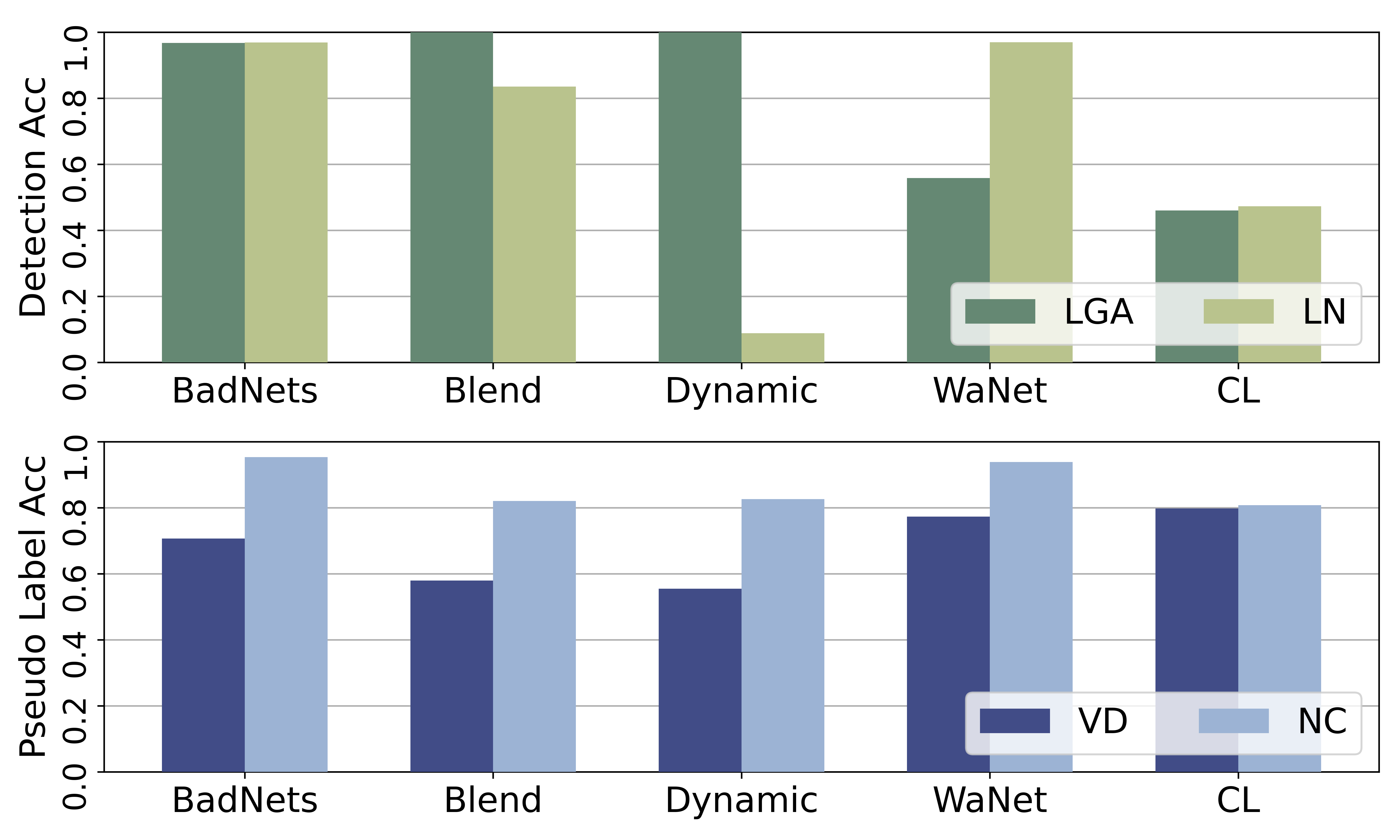}
    \caption{Detection and pseudo label accuracy on CIFAR-10. The maximal detection accuracy for CL attack is $\min(\frac{\lambda}{\mu}, 1)=0.5$. Pseudo label accuracy is calculated on LGA detected samples.}
    \label{fig:detection_label_acc} 
\end{figure}

\subsection{Main Results}
\paragraph{Backdoor Detection and Pseudo Labels.} LGA is adopted for backdoor detection in this section. As shown in \cref{fig:detection_label_acc}, detection accuracy approaches its maximal value in most cases. However, the method has less satisfying performance under WaNet attack, which adopts a noise mode to escape detection.
We generate pseudo labels with VD and NC. \cref{fig:detection_label_acc} shows that the latter method based on self-supervised learning generates pseudo labels of higher quality.
In some cases, NC achieves very high accuracy on detected samples. We attribute this partially to the fact that LGA prefers to isolate images whose losses drop faster. These samples have more salient class features and are thus closer to the corresponding centers. 
On tiny-ImageNet, detection accuracy approaches 100\%, but the label accuracy of VD is 46.64\%, 38.90\%, 42.34\% for BadNets, Blend, WaNet respectively, posing a greater challenge than CIFAR-10. 

\paragraph{Comparison with NAD and ABL.} 
As shown in \cref{tab:main_cifar10} and \cref{tab:main_tiny}, NAB outperforms NAD and ABL by a large margin across all the settings in terms of clean accuracy. CA of our method is lower under WaNet and CL attack than other attacks because the a large number of clean samples are incorporated into the detection data $\mathcal{D}_s'$, but still higher than that of the baseline defenses.
Our method also has outstanding performance in terms of attack success rate. It obtains the lowest ASR in most cases. On tiny-ImageNet, however, ABL achieves a significantly low ASR. We suspect that the diverse classes of the dataset help the unlearning stage of ABL identify backdoor features more precisely.
We also find that results of our method are even better on ResNet-50, achieving a much lower ASR and a clean accuracy comparable to \textit{no defense} in some cases. Model capacity might benefit the injection of an additional backdoor.
In summary, our method suppresses the attacker's backdoor effectively while having limited influence on clean accuracy. By simply poisoning part of the training data, our method achieves state-of-the-art defense performance.

\paragraph{Comparison with DBD.} DBD adopts self-supervised learning (SSL) in its first training stage, and part of the impressive performance on clean accuracy comes from the extra training epochs. We also use the SSL pretrained network in pseudo label prediction (NC) and model initialization for fair comparison.
As shown in \cref{tab:main_ssl}, our method outperforms DBD by a large margin in both ASR and CA. NAB achieves better results even without SSL pre-training.
We also find that higher pseudo label accuracy obtained by NC helps reduce the performance drop on clean data. More analyses of this factor are presented in \cref{sec:further_ana}.

\begin{figure*} [t!]
\centering
    \subfloat[Clean Accuracy\label{fig:je_ca}]{
        \includegraphics[width=0.5\linewidth]{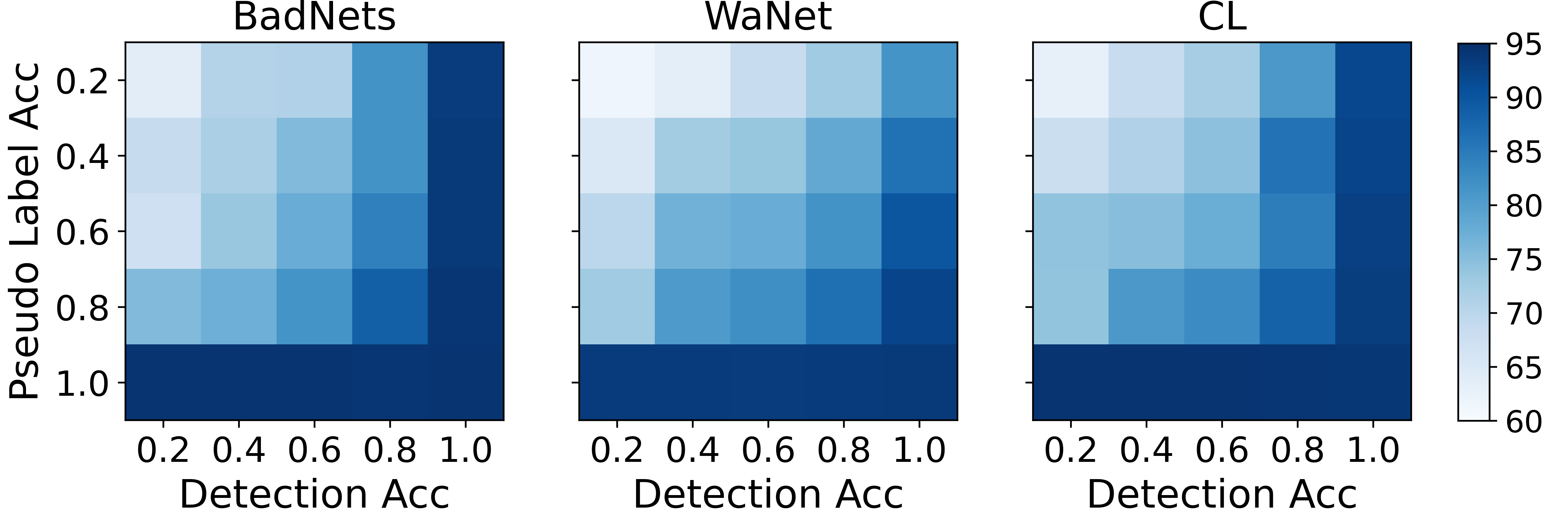}
    }
    \subfloat[Backdoor Accuracy\label{fig:je_ba}]{
        \includegraphics[width=0.5\linewidth]{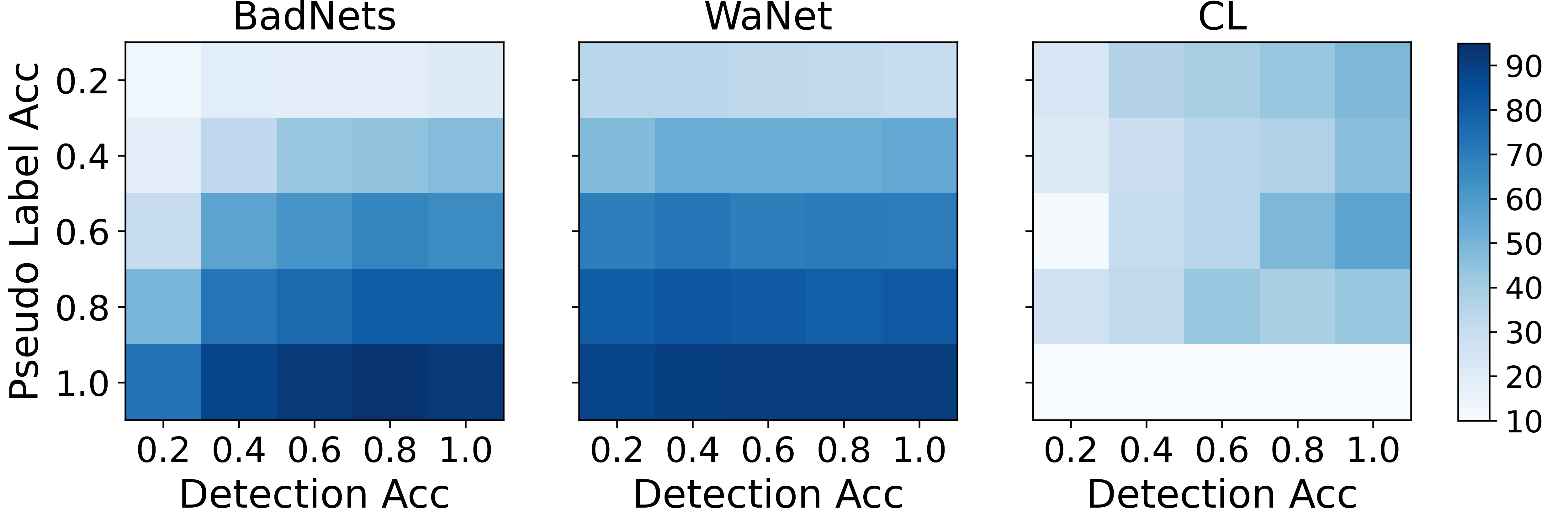}
    }
    \\
    \vspace{-0.6em}
    \subfloat[Attack Success Rate\label{fig:je_asr}]{
        \includegraphics[width=0.5\linewidth]{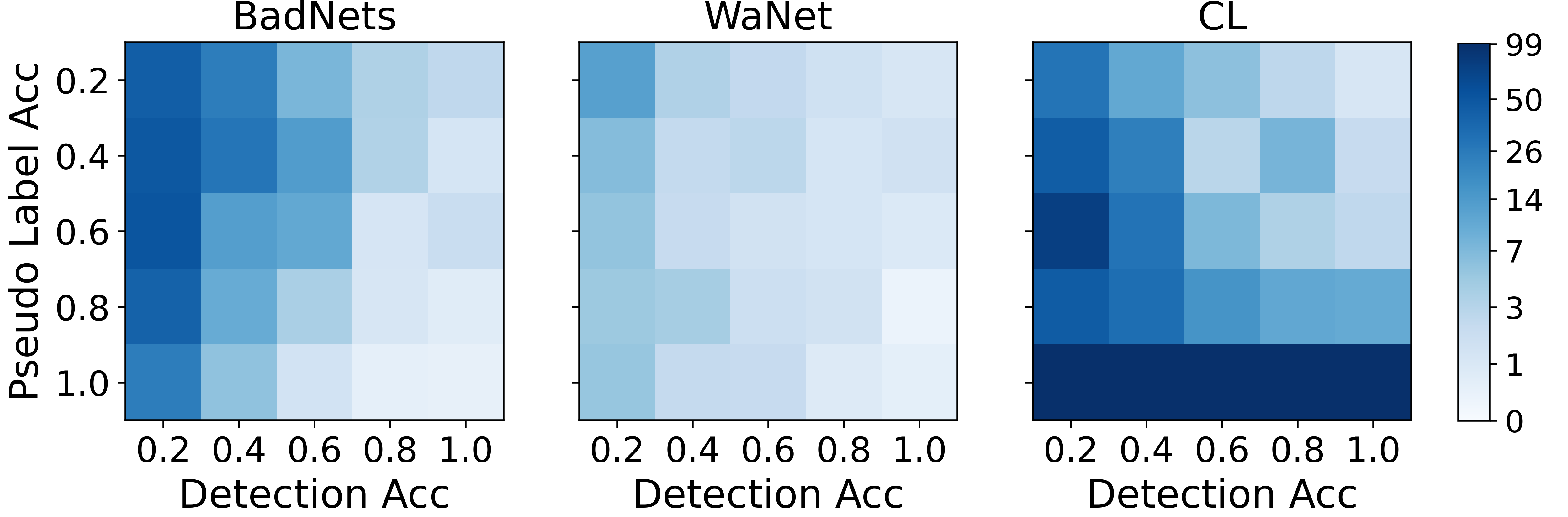}
    }
    \subfloat[Defense Success Rate\label{fig:je_dsr}]{
        \includegraphics[width=0.5\linewidth]{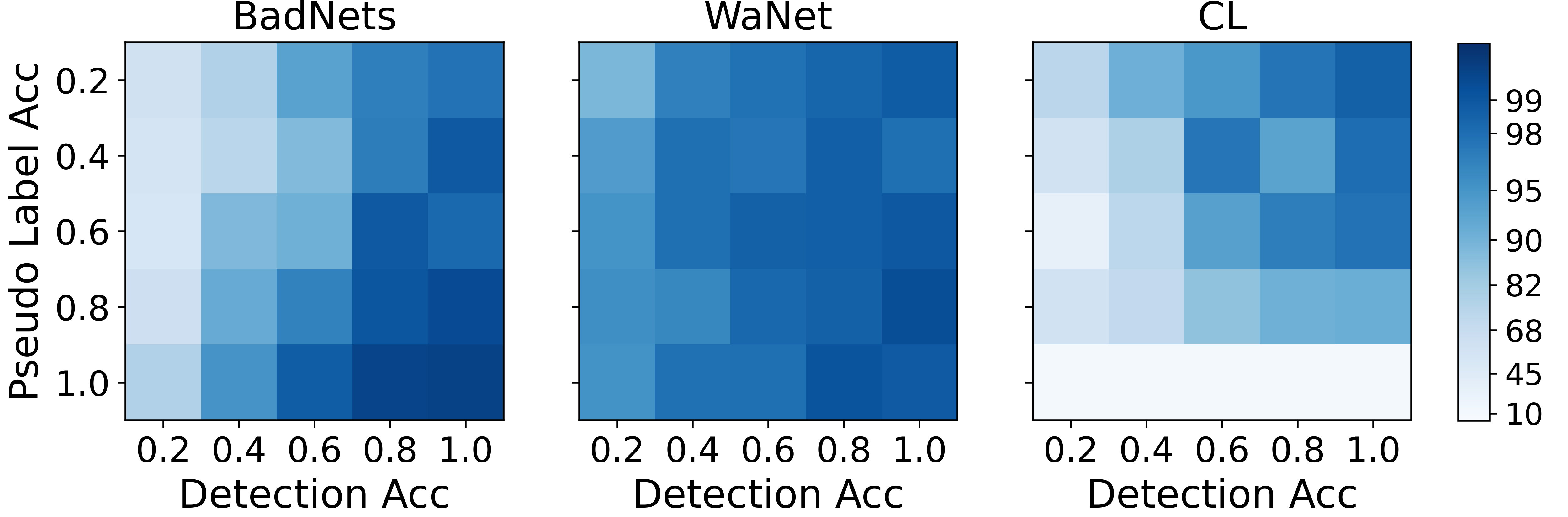}
    }
    \setlength{\abovecaptionskip}{0.2em}
    \caption{Clean accuracy, backdoor accuracy, attack success rate and defense success rate (\%) under different detection accuracy and pseudo label accuracy. The experiments are conducted on CIFAR-10 under BadNets, WaNet and CL. To generate pseudo labels of accuracy $p$, we randomly change $1-p$ of the true labels to a different class. For a detection accuracy $q$, we randomly select $qN$ poisoned samples and $(1-q)N$ clean samples, where $N$ is size of the training set.}
    \label{fig:je_all} 
\end{figure*}

\subsection{Effectiveness of Data Filtering}
We validate the effectiveness of data filtering and present the results in \cref{tab:filter}.
For all the defenses listed, accuracy on poisoned data lags behind that on clean data. The gap is especially obvious for NAB since the model learns to predict a stamped sample to its pseudo label which is typically not very accurate.
Augmenting NAB with data filtering provides a remedy for this.
We find that the defense success rate reaches over 99\% in all cases, suggesting that the filtering technique identifies most poisoned samples and those escaping from detection are typically correctly classified. Part of the clean samples are also rejected, but a significant performance drop is not observed. This is because most of the rejected clean samples are also misclassified. 

\begin{table}
    \centering
    \setlength\tabcolsep{1.6pt}
    \renewcommand{\arraystretch}{1.15}
    \begin{tabular}{c|cc|cc|cc|cc}
    \toprule[1.2pt]
        {\textbf{$\lambda$}} & \multicolumn{2}{c|}{\textbf{0.01}} & \multicolumn{2}{c|}{\textbf{0.05}} & \multicolumn{2}{c|}{\textbf{0.10}} & \multicolumn{2}{c}{\textbf{0.20}} \\\hline
        \textbf{Metric} & CA & ASR & CA & ASR & CA & ASR & CA & ASR  \\\hline
        \textbf{$\mu\downarrow$} & \multicolumn{8}{c}{\textbf{BadNets}} \\\hline
        \textbf{0.00} & 94.55 & 100 & 94.39 & 100 & 93.96 & 100 & 94.00 & 100 \\
        \textbf{0.01} & 94.45 & 1.00 & 93.97 & 8.76 & 94.18 & 61.96 & 93.91 & 78.72 \\
        \textbf{0.05} & 94.45 & 0.77 & 93.96 & 0.61 & 93.29 & 1.19 & 93.22 & 1.29 \\
        \textbf{0.10} & 91.23 & 0.79 & 92.81 & 0.67 & 93.01 & 0.13 & 93.01 & 0.22 \\\hline
        \textbf{$\mu\downarrow$} & \multicolumn{8}{c}{\textbf{Dynamic}} \\\hline
        \textbf{0.00} & 94.71 & 99.64 & 94.38 & 99.99 & 94.41 & 99.99 & 94.22 & 99.99 \\
        \textbf{0.01} & 94.34 & 1.06 & 94.19 & 28.99 & 94.05 & 53.71 & 93.70 & 82.17 \\
        \textbf{0.05} & 93.61 & 0.82 & 94.19 & 0.39 & 93.95 & 0.46 & 93.82 & 0.64 \\
        \textbf{0.10} & 90.66 & 0.87 & 91.98 & 0.42 & 93.16 & 0.24 & 93.69 & 0.31 \\
        
    \bottomrule[1.2pt]
    \end{tabular}
    \caption{Defense effectiveness (\%) of out method under different detection rate $\mu$ and poisoning rate $\lambda$ on CIFAR-10. SPECTRE and NC are adopted for detection and pseudo label generation, respectively.}
    \label{tab:ana_rate}
\end{table}

\subsection{Further Analyses}
\label{sec:further_ana}
\paragraph{Detection and Pseudo Label Accuracy.} Backdoor detection and pseudo label generation are two major components influencing the performance of our method. Analyses on them helps understand the robustness of NAB and can guide the selection of specific detection and relabeling strategies.
Therefore, we test NAB under different detection accuracy (DA) and pseudo label accuracy (PLA), and present the results in \cref{fig:je_all}. 
The following patterns can be observed: 
1) Clean accuracy (CA) relies on both DA and PLA, and NAB can preserve a high CA when either DA or PLA reaches a high level. 
2) Attack success rate (ASR) is more sensitive to DA, as the metric directly influences how many poisoned samples are stamped for non-adversarial backdoor injection. 
3) Backdoor accuracy (BA) mainly depends on PLA, but defense success rate (DSR) is more sensitive to DA. 
In practice, it is typically easy to find a detection method with accuracy over 90\%, but pseudo label accuracy varies in different scenarios. 
CL, which is representative of label-consistent attacks, shows different reactions to PLA. 
The attack does not change the labels of poisoned samples, so actually we need incorrect pseudo labels to break the backdoor correlation. 
It is also worth mentioning that accuracy cannot fully reflect the quality of backdoor detection and pseudo label generation. 
For example, a detection method might have \textit{detection bias}, which means it has a preference for detecting poisoned samples with some explicit patterns. A strong detection bias might hamper the defense performance of NAB even under a high DA. We leave a discussion on this problem in the supplementary material.

\paragraph{Detection Rate and Poisoning Rate.}
The defender needs to specify the detection rate $\mu$
 without being aware of the poisoning rate $\lambda$. We experiment with different $\mu$ and $\lambda$ and display the results in \cref{tab:ana_rate}. When $\mu > \lambda$, the detection accuracy drops below $\frac{\lambda}{\mu}$. Performance on clean data is hampered to some extent, which is consistent with the conclusion made in the previous paragraph. 
 When $\mu \leq \lambda$, NAB demonstrates satisfying performance on both CA and ASR. However, the defense effectiveness decays significantly when we have $\mu \ll \lambda$. The injected non-adversarial backdoor is not strong enough to suppress the attacker's backdoor in this case.
 Typically, attackers choose a small $\lambda$ to escape inspection. Our choice $\mu=0.05$ in the main experiments suffices to handle most situations.

\paragraph{Effectiveness under All-to-All Attack.}
So far we have tested NAB under all-to-one attack (A2O), where all the poisoned samples are relabeled to a single target class. 
In this section we introduce all-to-all attack (A2A) where samples with different original labels have different target labels. 
As shown in \cref{tab:ana_a2a}, A2A is less effective than A2O in terms of ASR. Our method can successfully defend against A2A, but the defense effectiveness is slightly lower than under A2O. 
Besides, it can be found that model capacity brings more benefits under A2A. We attribute it to that larger networks provide more learning ability to handle the complexified tasks in \cref{eq:two_tasks} and \cref{eq:defense_task}.

\begin{table}[t!]
    \centering
    \setlength\tabcolsep{3.8pt}
    \renewcommand{\arraystretch}{1.15}
    \begin{tabular}{c|c|cc|cc}
    \toprule[1.2pt]
        \multicolumn{2}{c|}{\textbf{Attack}} & \multicolumn{2}{c|}{\textbf{BadNets}} & \multicolumn{2}{c}{\textbf{Blend}}\\\hline
        \textbf{Defense $\downarrow$} & \textbf{Arch $\downarrow$} & CA & ASR & CA & ASR  \\\hline
        \multirow{2}{*}{\textbf{None}} & ResNet-18 & 94.29 & 93.37 & 93.75 & 90.12\\
         & ResNet-50 & 94.53 & 93.97 & 94.20 & 90.87\\\hline
        \multirow{2}{*}{\textbf{NAB}} & ResNet-18 & 93.24 & 2.66 & 93.28 & 1.48\\
         & ResNet-50 & 94.36 & 1.28 & 94.61 & 1.19\\
    \bottomrule[1.2pt]
    \end{tabular}
    \caption{Attack effectiveness (\%) of all-to-all attacks and defense effectiveness (\%) of our method under them on CIFAR-10. LN and NC are adopted for detection and pseudo label generation respectively.}
    \label{tab:ana_a2a}
\end{table}

\subsection{Understanding Non-Adversarial Backdoor}
To get a comprehensive understanding of how NAB works, we first visualize the saliancy maps to illustrate how much attention the models pay on particular area of the input images. 
As shown in \cref{fig:saliency}, the stamp (a $2\times 2$ patch on the upper left corner) catches much attention when the trigger pattern is also added, but has a much weaker influence on clean data. 
This is consistent with the observation that the stamp can greatly change the behavior of a model only when the inputs are poisoned.
We also visualize the representations in \cref{fig:representations} to have a deeper insight into the mechanism behind our defense. 
Representations of stamped samples and clean samples are mixed up together, while those of poisoned samples are clearly separated except on the target class. 
These findings further demonstrate that our defense does not actually mitigate the attacker's backdoor, but inject a non-adversarial backdoor to suppress it. 
Besides, the model can directly predict the authentic labels of poisoned samples given a set of accurate pseudo labels.

\section{Future Exploration with NAB Framework}
We stress that the value of NAB is not limited to its simplicity, flexibility and impressive performance.
The framework introduces the idea of backdoor for defense, which we believe is worthy of further exploration just like backdoor attack.
Our implementation of NAB is not claimed to be optimal, and a lot more efforts can be done to strengthen it.
We only list a few directions due to page limitation:

\begin{figure}[t!]
    \centering
    \setlength{\abovecaptionskip}{0.3em}
    \includegraphics[width=0.92\linewidth]{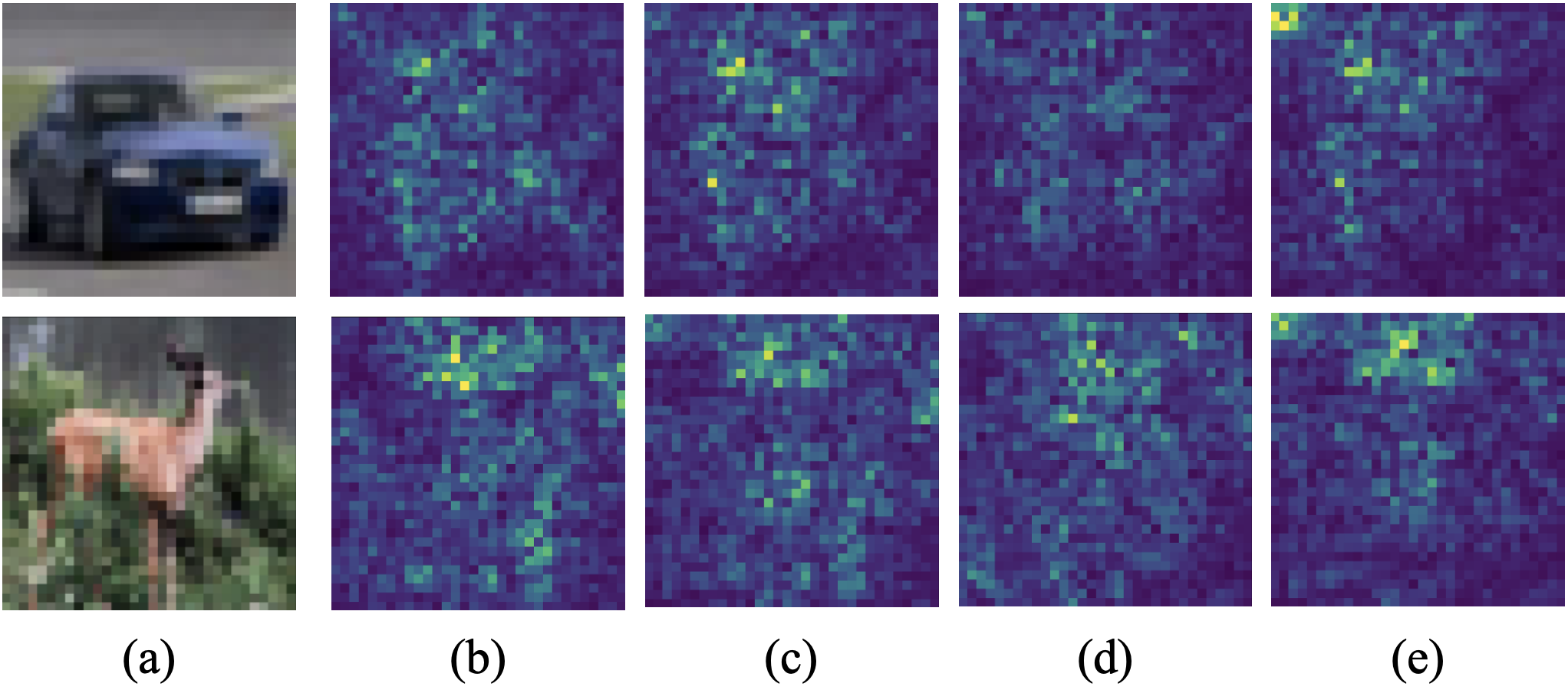}
    \caption{Examples of (a) raw images and saliency maps of their (b) clean, (c) stamped clean, (d) poisoned, (e) stamped and poisoned versions, which are obtained using NAB under BadNets attack.}
    \label{fig:saliency} 
\end{figure}

\paragraph{Protection for clean samples.}
The detection and relabeling accuracy might be both low in some cases. The non-adversarial backdoor would then be triggered on some clean samples and brings performance drop. The problem will possibly be alleviated by providing a protection mechanism (\textit{e.g.} stamping some clean stamps without relabeling).

\paragraph{Sample-efficient backdoor.} Injecting backdoor in a sample-efficient way has been a hot topic in backdoor attack \cite{xia2022data, zeng2022narcissus}. The defender will also want to inject a backdoor strong enough for defense with as few samples as possible. A benefit of sample-efficient backdoor for NAB is that when the number of required samples is small enough, the detected samples can go through human inspection and relabeling, guaranteeing a high DA and PLA.

\paragraph{Backdoor vaccination.} A even more interesting question is whether the defender can carry out a backdoor attack, defend against it using NAB, and generalize the defense effectiveness to other attacks. 
We test the idea under a quite limited setting where the target class is known. 
The results displayed in the supplementary material show that ASR of Blend and WaNet attack is to some extent hampered by the defense targeting BadNets. If the generalization ability of defender’s backdoor is further improved, NAB can dispose of backdoor detection and pseudo label generation since it only needs to focus on the attack transparent to defender.

\section{Conclusion}
In this work, we propose a novel defense framework NAB which injects a non-adversarial backdoor targeting poisoned data. 
Following the procedures in backdoor attack, we detect a small set of suspicious samples and process them with a poisoning strategy.
During inference, we keep the non-adversarial backdoor triggered to suppress the effectiveness of attacker's backdoor.
Extensive experiments demonstrate that NAB can achieve successful defense with minor performance drop on clean data.
NAB has long-term value both as a powerful defense method and as a potential research area.
As a method, its components are highly replaceable and can be updated and optimized in the future. 
As a research area, we hope that stronger variants would be derived from the simple and flexible framework, just as what has happened in backdoor attack.

{\small
\bibliographystyle{ieee_fullname}
\bibliography{ref}
}


\end{document}